\pdfoutput=1
\documentclass[11pt, letterpaper, logo, onecolumn, copyright, numbering]{paper_improved}

\usepackage{soul}
\usepackage{xcolor}

\usepackage[export]{adjustbox} 
\usepackage{enumitem}
\usepackage{array}
\usepackage{colortbl}
\usepackage{xcolor}
\usepackage{caption}
\usepackage{tabularray}
\usepackage[utf8]{inputenc} 
\usepackage[T1]{fontenc}    
\usepackage{hyperref}       
\usepackage{url}            
\usepackage{booktabs}       
\usepackage{amsfonts}       
\usepackage{nicefrac}       
\usepackage{microtype}      
\usepackage{xcolor}         
\usepackage{graphicx}

\usepackage{natbib}
\usepackage{verbatim}
\usepackage{inconsolata}
\usepackage{tabularx}
\usepackage{float}
\usepackage{multirow}
\usepackage{listings}
\usepackage{xcolor}
\usepackage{subfigure}
\usepackage{subcaption}
\usepackage{amsmath}
\usepackage{amssymb}
\usepackage{array}
\usepackage{svg}
\usepackage[T1]{fontenc}
\usepackage{wrapfig}

\definecolor{lightgray}{rgb}{0.95, 0.95, 0.95}
\definecolor{darkgray}{rgb}{0.4, 0.4, 0.4}
\definecolor{backcolour}{rgb}{0.95,0.95,0.92}
\definecolor{myblue}{rgb}{0.2, 0.4, 0.8} 
\definecolor{mygreen}{rgb}{0.2, 0.6, 0.2} 

\usepackage{amsmath,amssymb,amsfonts}

\usepackage{CJKutf8}
\usepackage{soul}
\usepackage{url}
\usepackage[utf8]{inputenc}
\usepackage{caption}
\usepackage{xcolor}
\usepackage{amsmath}

\usepackage{amsthm}
\usepackage{booktabs}
\usepackage{algorithm}
\usepackage{latexsym}
\usepackage{amsmath}
\usepackage{subfigure}
\usepackage{microtype}
\usepackage{algorithmic}
\usepackage[switch]{lineno}
\usepackage[most]{tcolorbox}
\usepackage{alltt}

\definecolor{forestgreen}{rgb}{0.13, 0.55, 0.13}

\tcbset{
    aibox/.style={
        colback=white, 
        colframe=blue!75!black, 
        colbacktitle=blue!85!black, 
        coltitle=white, 
        fonttitle=\bfseries,
        enhanced,  
        drop shadow=black!50!white,  
        boxrule=1pt,  
        boxsep=10pt,  
        left=10pt,  
        right=10pt,  
        top=6pt,  
        bottom=6pt,  
        title code={\path[tcb fill frame] ([xshift=-10pt]frame.west) -- (frame.north west) -- (frame.north east) -- ([xshift=10pt]frame.east) -- cycle;},  
        attach boxed title to top left={xshift=10pt, yshift*=-\tcboxedtitleheight/2},
        boxed title style={boxrule=0pt, frame code={}}  
    }
}

\newtcolorbox{AIbox}[2][]{aibox, title=#2, #1}

\let\cite\citep

\title{DeepWideSearch: Benchmarking Depth and Width in Agentic Information Seeking}
\author[*,1]{Tian Lan, Bin Zhu, Qianghuai Jia, Junyang Ren, Haijun Li, Longyue Wang{$^*$}, Zhao Xu, Weihua Luo, Kaifu Zhang\\ \bf Alibaba International Digital Commerce \\
\vspace{1mm} $*$ Corresponding Author: Longyue Wang\\
}

\begin{abstract}
{\bf \large Abstract}\vspace{1mm}

Current search agents fundamentally lack the ability to simultaneously perform \textit{deep} reasoning over multi-hop retrieval and \textit{wide}-scale information collection—a critical deficiency for real-world applications like comprehensive market analysis and business development.
To bridge this gap, we introduce DeepWideSearch, the first benchmark explicitly designed to evaluate agents to integrate depth and width in information seeking. In DeepWideSearch, agents must process a large volume of data, each requiring deep reasoning over multi-hop retrieval paths.
Specifically, we propose two methods to converse established datasets, resulting in a curated collection of 220 questions spanning 15 diverse domains. Extensive experiments demonstrate that even state-of-the-art agents achieve only 2.39\% average success rate on DeepWideSearch, highlighting the substantial challenge of integrating depth and width search in information-seeking tasks. 
Furthermore, our error analysis reveals four failure modes: lack of reflection, overreliance on internal knowledge, insufficient retrieval, and context overflow—exposing key limitations in current agent architectures.
We publicly release DeepWideSearch to catalyze future research on more capable and robust information-seeking agents.
\\
\\
\makebox[1pt][l]{\parbox{\textwidth}{\raggedright\begin{tabular}{@{} l l @{}} \raisebox{-0.5em}{\includegraphics[height=1.6em]{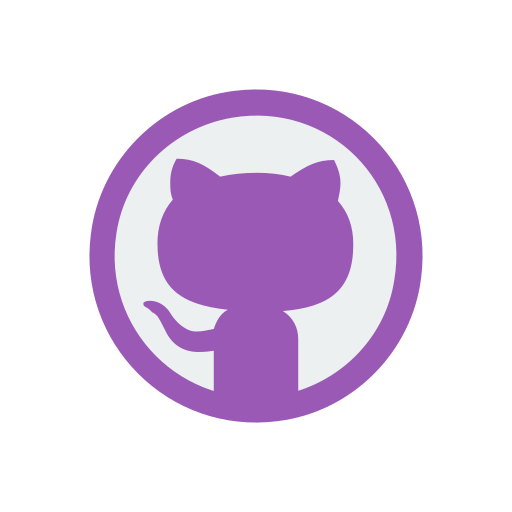}} & \small\url{https://github.com/AIDC-AI/Marco-Search-Agent}\\ \raisebox{-0.3em}{\includegraphics[height=1.4em]{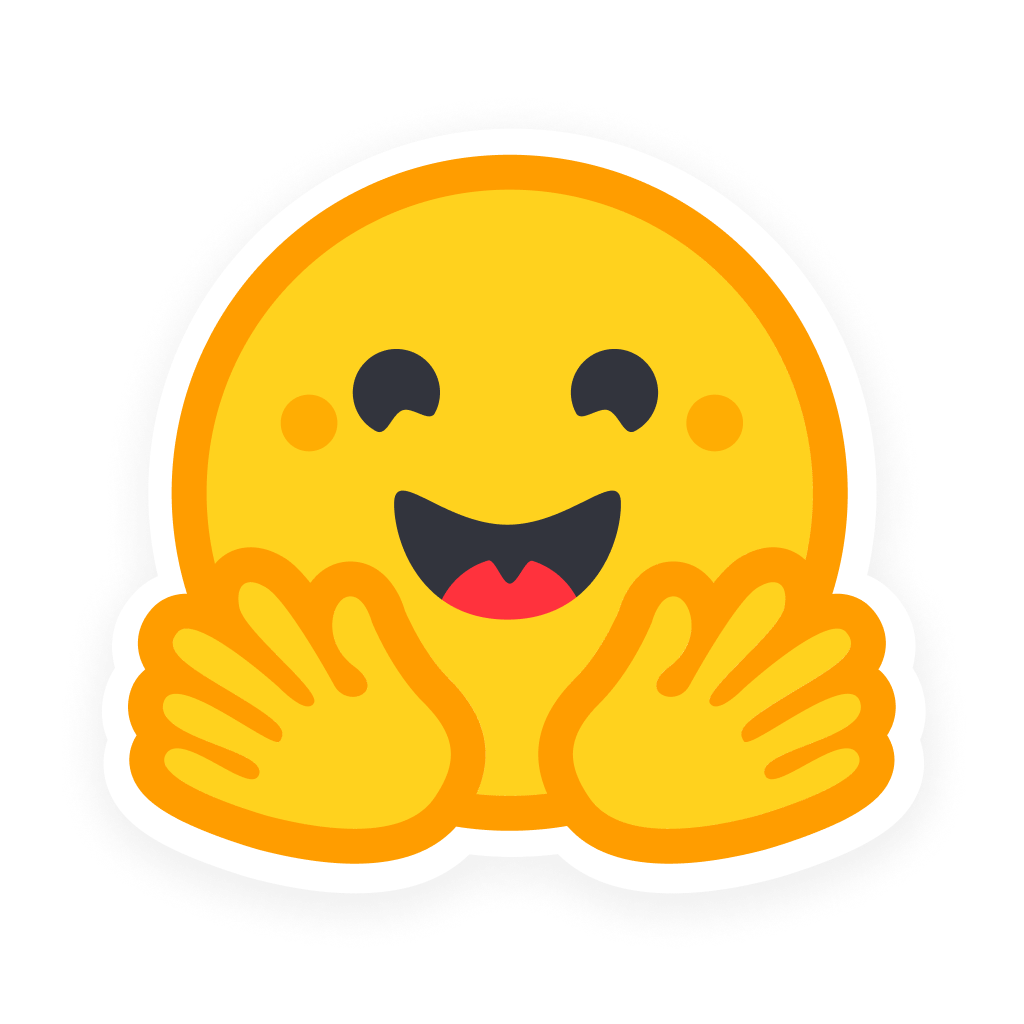}} &\small\url{https://huggingface.co/datasets/AIDC-AI/DeepWideSearch} \end{tabular}}}
\end{abstract}

\begin{document}

\maketitle

\section{Introduction}

\begin{wrapfigure}[18]{r}{0.425\textwidth} 
\vspace{-25pt}
  \centering
  \raisebox{-0.5\height}{\includegraphics[width=0.425\textwidth]{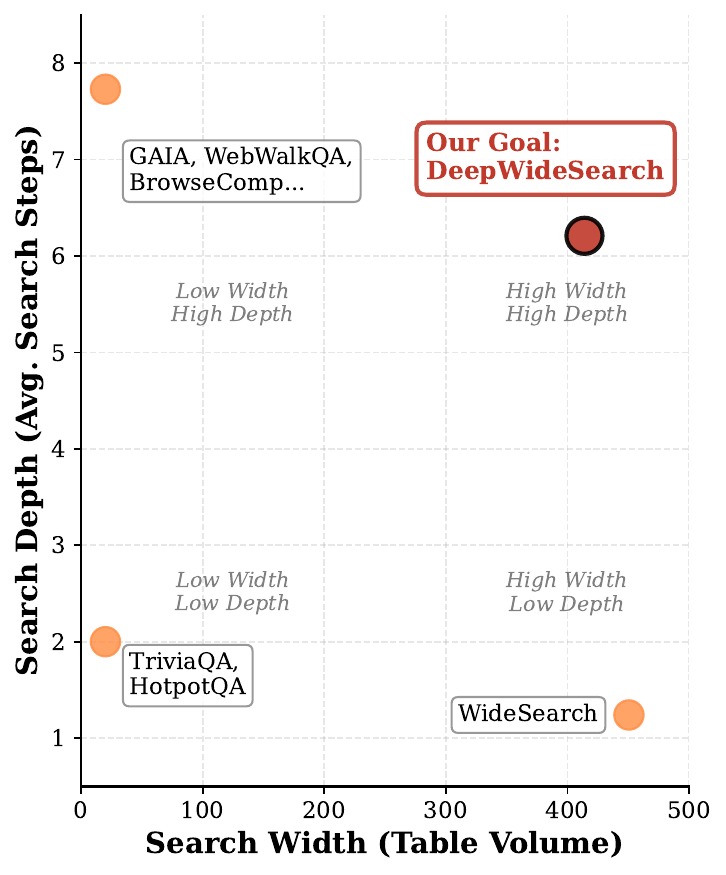}}
  \vspace{-10pt}
  \captionof{figure}{The comparison of existing benchmarks on search width and depth.}
  \label{img:teaser_img}
\end{wrapfigure}

Large Language Models (LLMs) with advanced reasoning capabilities~\cite{achiam2023gpt,liu2024deepseek,guo2025deepseek} have driven substantial progress across a wide range of natural language tasks. Building on these advances, LLM-based agents that equipped with planning, tool use, and multi-step reasoning capabilities~\cite{xi2025survey,gao2025survey}—have achieved strong performance on complex real-world challenges, including computer operation~\cite{wang2025opencuaopenfoundationscomputeruse}, deep research~\cite{du2025deepresearchbenchcomprehensivebenchmark}, and  information seeking~\cite{mialon2023gaiabenchmarkgeneralai,wei2025browsecompsimplechallengingbenchmark}.


\begin{figure}[h]
  \centering
  \includegraphics[width=\textwidth]{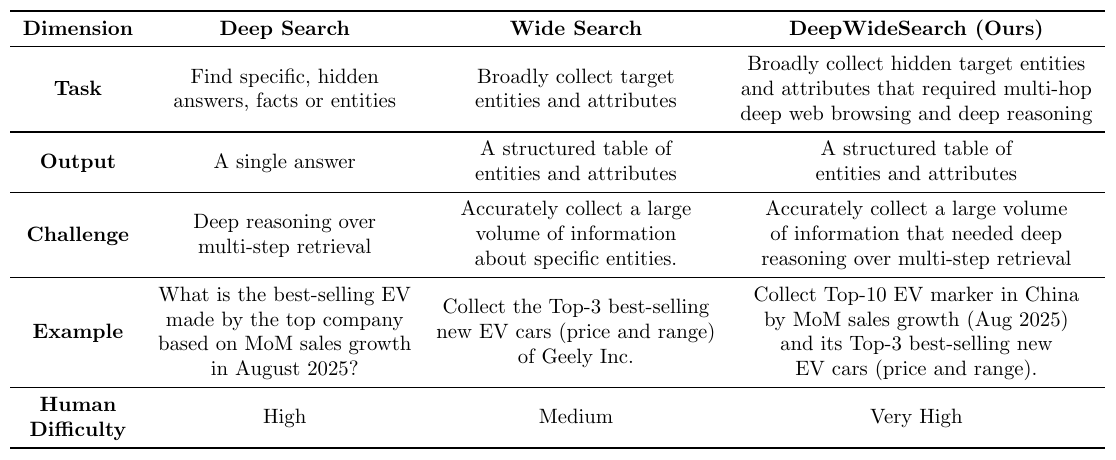}
  \caption{The detailed comparison among deep search, wide search benchmarks and our proposed DeepWideSearch.}
  \label{img:teaser_table}
  \vspace{-15pt}
\end{figure}

5o far, existing benchmarks for evaluating agents can be systematically categorized along two critical dimensions (Figure~\ref{img:teaser_img}): search width (measured by the number of information units to be searched) and search depth (measured by average search steps for each unit), revealing four distinct categories: (1) \textit{Low width, high depth} benchmarks (e.g., GAIA~\cite{mialon2023gaiabenchmarkgeneralai}, BrowseComp~\cite{wei2025browsecompsimplechallengingbenchmark}), which focus on intricate deep reasoning over multi-hop retrieval for searching target answers; (2) \textit{Low width, low depth} benchmarks (e.g., TriviaQA, HotpotQA), which address simple fact-finding tasks; (3) \textit{High width, low depth} benchmarks (e.g., WideSearch~\cite{wong2025widesearch} and PaSa~\cite{he2025pasa}), which emphasize broad information collection about specific questions; and critically, (4) \textit{High width, high depth} tasks, which collect extensive information that required deep reasoning—a critical capability for real-world applications like comprehensive market analysis and business development but entirely unaddressed by current benchmarks. 
For instance, as shown in Figure~\ref{img:teaser_table}, the case \textit{``identifying the Top-10 EV maker in China by MoM sales growth (Aug 2025) and its Top-3 best-selling new EV cars (price and range)''} exemplifies this challenge.
It requires agent to gather a large volume of candidates, \textit{i.e.}, \textit{EV makers}, to fill the result table through wide-scale search, and verify each candidate by performing deep reasoning, a combinatorial complexity that exceeds both the scope of width-focused evaluations and the scale of depth-focused benchmarks.

To address this critical evaluation gap, we introduce DeepWideSearch, the first benchmark explicitly designed to evaluate the capability of agents in deep and wide information seeking. 
Since it is challenging to construct deep and wide search instances even with human annotation, we develop two methods for conversing established datasets: (1) \textit{Deep2Wide Conversion}, which extends deep search benchmarks (e.g., GAIA and BrowseComp) by augmenting their information scope through human-annotated table schemas; and (2) \textit{Wide2Deep Conversion}, which enhances wide search queries by replacing explicit entities with synthesized complex sub-questions that necessitate multi-hop search steps. Both approaches integrate rigorous human validation protocols to ensure data quality while maintaining the combinatorial complexity inherent in real-world information-seeking scenarios. The final benchmark comprises 220 meticulously curated questions spanning 15 diverse domains, featuring both Chinese and English queries with human-verified ground truths, with 85 instances derived from Deep2Wide and 135 from Wide2Deep construction methods.

We conduct comprehensive experiments across state-of-the-art LLMs and agent systems on DeepWideSearch. Our results demonstrate that even the most advanced agent systems achieve only 2.39\% average success rate on DeepWideSearch, highlighting the substantial difficulty of this kind of information seeking task. Notably, while agent frameworks consistently improve core entity identification (e.g., +15.91 absolute percentage points in Core Entity Accuracy), they exhibit limited efficacy in wide-scale information collection, frequently underperforming their LLMs counterparts using internal knowledge. Through systematic error analysis, we identify four fundamental failure modes: (1) lack of effective reflection mechanisms when encountering problematic search trajectories; (2) overreliance on parametric internal knowledge leading to outdated or inaccurate information; (3) insufficient retrieval despite accessing relevant webpages; and (4) context overflow exceeding current agent architectue limitations. These empirical findings expose key limitations of current agent architecture for the deep and wide information-seeking task. To facilitate further research in this critical domain, we have publicly released the DeepWideSearch benchmark, including datasets and evaluation codebase.

\section{Related Work}

\subsection{LLM-based Search Agents}
The emergence of LLM-based agent systems has enabled sophisticated information-seeking capabilities, with frameworks ranging from closed-source implementations (e.g., OpenAI Deep Research) to open-source platforms (e.g., WebAgent~\cite{wu2025webwalkerbenchmarkingllmsweb} and Cognitive Kernel-Pro~\cite{fang2025cognitivekernelproframeworkdeep}). These systems have demonstrated proficiency in numerous application domains, including computer-use agents, deep research for complex problem investigation~\cite{han2025deepresearchertesttimediffusion}, and multi-step information retrieval through tool use~\cite{xi2025survey}. Among these applications, information-seeking agents represent a critical frontier impact real-world utility.
Current research in this domain primarily addresses five technical challenges: (1) agentic system architecture design~\cite{zhang2025agentorchestra,zhou2025multiagentdesignoptimizingagents,xia-etal-2025-agentrm,fang2025webevolverenhancingwebagent}, (2) synthetic data generation for complex scenarios~\cite{wu2025webdancerautonomousinformationseeking,li2025websailor,tao2025webshaper}, (3) optimization techniques for retrieval efficiency~\cite{zhang2025rlvmrreinforcementlearningverifiable,fan2025ssrlselfsearchreinforcementlearning,sun2025zerosearchincentivizesearchcapability}, (4) knowledge management for multi-hop reasoning~\cite{zhang2025agentorchestra,xu2025amemagenticmemoryllm}, and (5) evaluation methodologies for performance assessment~\cite{zhuge2025agentasajudge,gou2025mind2web2evaluatingagentic}.

\subsection{Benchmarks for LLM-based Agents}
Existing evaluation frameworks for information-seeking agents primarily target two distinct capabilities: (1) \textit{Depth} in multi-hop reasoning, measured by benchmarks like GAIA~\cite{mialon2023gaiabenchmarkgeneralai} and BrowseComp~\cite{wei2025browsecompsimplechallengingbenchmark} for general complex reasoning, and domain-specific variants in healthcare~\cite{chen2025medbrowsecompbenchmarkingmedicaldeep} and E-commerce~\cite{lyu2025deepshopbenchmarkdeepresearch}; (2) \textit{Width} in information collection, assessed by WideSearch~\cite{wong2025widesearch} for comprehensive retrieval of atomic information, and PaSa~\cite{he2025pasa} and SPAR~\cite{shi2025sparscholarpaperretrieval} for academic literature retrieval.
Crucially, no existing benchmark captures the \textit{combinatorial complexity} inherent in real-world information-seeking tasks that simultaneously demand extensive exploration (width) and intricate multi-step reasoning (depth). This fundamental gap in evaluation methodology has prevented meaningful progress toward agents capable of handling the complex real-world information-seeking. To address this limitation, we propose DeepWideSearch, the first benchmark explicitly designed to evaluate the capability of agents in the deep and wide information-seeking task.

\section{Task Formulation}

\begin{figure*}[ht]
    \centering
    \includegraphics[width=\linewidth]{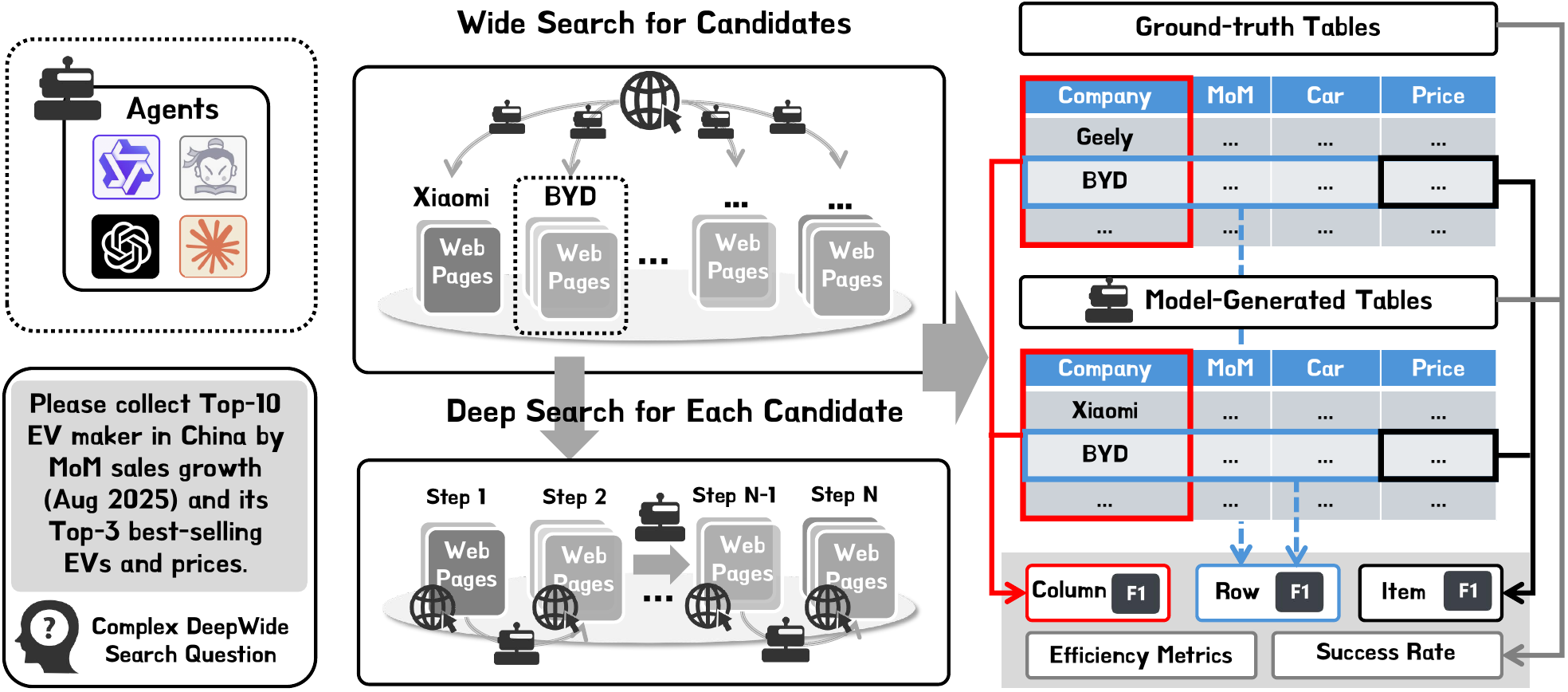}
    \caption{Task formulation of DeepWideSearch task. The evaluation metrics (highlighted in red) are detailed in Section~\ref{subsec.evaluation_metric}.}
    \label{fig:intro_case_illustration}
\end{figure*}

As shown in Figure~\ref{fig:intro_case_illustration}, DeepWideSearch establishes an evaluation framework that explicitly captures the \textit{combinatorial complexity} of real-world information-seeking tasks—requiring agents to perform \textit{deep} reasoning and \textit{wide}-scale information collection. 
The evaluation metrics (Column F1, Row F1, Item F1, and Success Rate) illustrated in Figure~\ref{fig:intro_case_illustration} will be formally described in Section~\ref{subsec.evaluation_metric}.

\paragraph{Input} Formally, each task in DeepWideSearch is defined as a tuple ($Q, C$):
(1) \textbf{Question $\boldsymbol{Q}$} represents a complex natural language query for deep and wide information seeking; 
and (2) \textbf{Columns $\boldsymbol{C=\{c_i\}_{i=1}^N}$} define the table schema as a set of $N$ attributes and constraints need to be collected and verified, such as EV price and MoM scales growth in Figure~\ref{fig:intro_case_illustration} (right).

\paragraph{Output} As shown in Figure~\ref{fig:intro_case_illustration} (medium), agents are required to generate a structured tabular response $R$ by performing wide search for gathering numerous candidates and deep search for the verification of each candidate.

\section{Methodology of Dataset Construction}

Constructing DeepWideSearch instances from scratch presents significant challenges due to the substantial human effort. To address this challenge while maintaining methodological rigor, we propose two methods to converse established datasets into deep and wide search questions: (1) Deep2Wide Conversion and (2) Wide2Deep Conversion. Both methodologies are complemented by human annotation procedures to ensure the quality.

\begin{figure*}[ht]
    \centering
    \includegraphics[width=\linewidth]{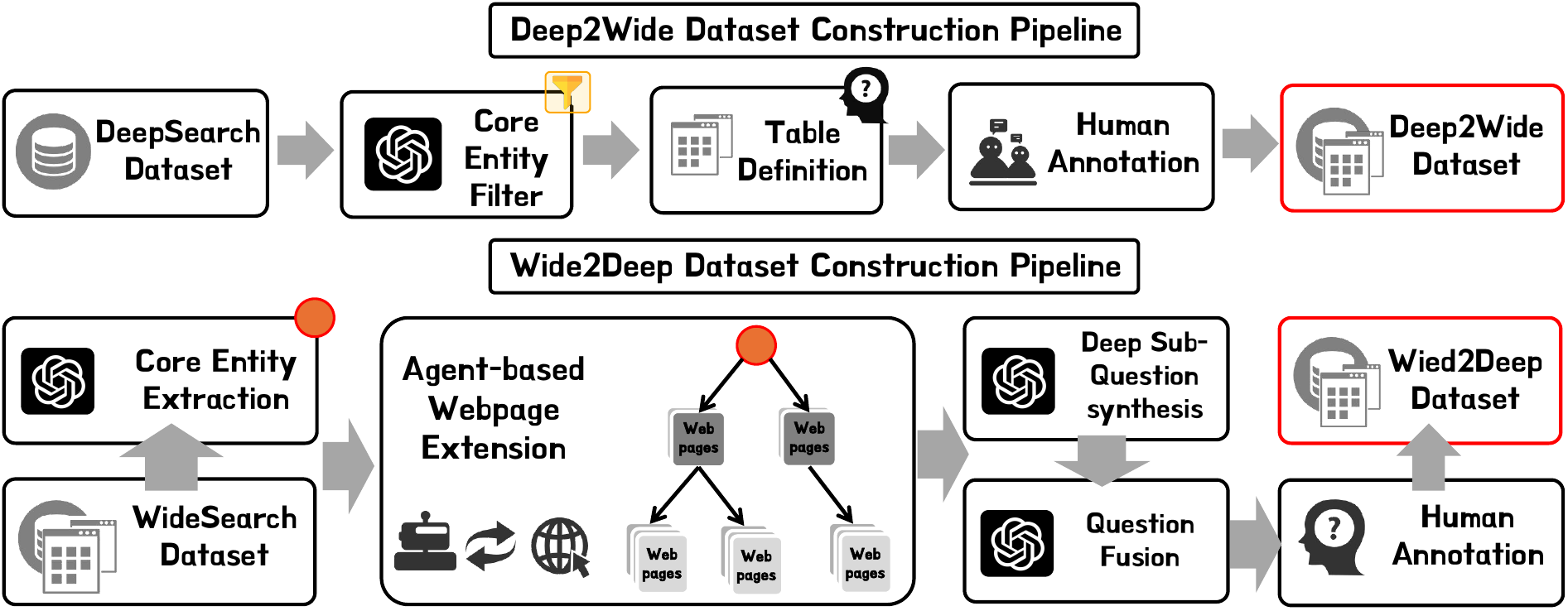}
    \caption{The pipelines of our proposed Deep2Wide and Wide2Deep data construction methods.}
    \label{fig:dataset_pipeline}
\end{figure*}

\subsection{Convert Deep Search Datasets (Deep2Wide)}

Existing deep search benchmarks such as GAIA~\cite{mialon2023gaiabenchmarkgeneralai}, BrowseComp~\cite{wei2025browsecompsimplechallengingbenchmark} and BrowseComp-zh~\cite{zhou2025browsecompzhbenchmarkingwebbrowsing} require agents to employ multi-hop web browsing and deep reasoning to identify target answers. Building upon these resources, we develop the Deep2Wide conversion methodology by expanding the scope of searched information. As illustrated in Figure~\ref{fig:dataset_pipeline} (Top), our approach follows a three-stage pipeline inspired by WideSearch \cite{wong2025widesearch}: (1) \textbf{Core Entity Filtering}: We sample 80 Chinese questions from BrowseComp-zh and 20 English questions from BrowseComp, filtering out instances where answers are unsuitable as core entities (\textit{e.g.}, dates and numerical values). For example, as shown in Figure~\ref{fig:lin_dan}, \textit{Dan Lin} is the core entity of the deep search question; 
(2) \textbf{Table Schema Definition}: Human annotators design structured table schemas by defining relevant information about the core entities; (3) \textbf{Comprehensive Annotation}: Annotators perform exhaustive web searches to populate the tables. Each instance requires approximately 30 minutes of human annotation time, ensuring high-quality and verified data.
Following a design similar to that of the WideSearch benchmark~\cite{wong2025widesearch}, we incorporated timestamps into each question to ensure that the answers remain invariant over time.

\begin{figure*}[htbp]
    \begin{tcolorbox}[
        colback=yellow!10,
        colbacktitle=black!80,
        coltitle=white,
        arc=1mm,
        boxrule=0pt,
        breakable]
        \textbf{Deep Search Question}: There is a Chinese athlete who has achieved remarkable success in a ball sport. He was the first player in his discipline to successfully defend his title at a major competition and has won multiple world championship titles. His sport underwent rule changes in the early 21st century, and he became the first men's singles Olympic champion under the new rules. \\
        \textbf{Core Entity}: Dan Lin.\\
        \textbf{Table Schema}: Competition records of Dan Lin from 2010 to 2020 including the following columns: Date, Tournament Name, Level, Event, Result, and Match Details (including opponent, score, and win/loss outcome).
    \end{tcolorbox}
    \caption{One deep search question in BrowseComp-ZH.\label{fig:lin_dan}}
\end{figure*}

\subsection{Convert Wide Search Datasets (Wide2Deep) \label{sec:convert_wide_search}}

Given that WideSearch \cite{wong2025widesearch} represents the publicly available dataset providing human-annotated tabular answers for wide-scale information-seeking, we develop the Wide2Deep conversion methodology to transform these wide search queries by introducing complexity in entity identification. This approach reuse the valuable human-annotated table in WideSearch while enhancing the deep reasoning requirements. Inspired by WebWalker \cite{wu2025webwalkerbenchmarkingllmsweb}, we implement a human-in-the-loop pipeline (Figure 3, bottom) comprising five stages: (1) \textbf{Entity Extraction}: Advanced LLMs identify core entities in 160 English and Chinese WideSearch questions, similar to the core entity in the deep search benchmark (Figure~\ref{fig:lin_dan}); 
(2) \textbf{Deep Sub-Question Synthesis}: Following prior work \cite{li2025websailor,tao2025webshaper}, a web search agent are implemented to recursively traverse official websites about core entities and collecting their rich entity information. Then, a complex sub-question is generated based on these rich information, adhering to two critical constraints: (a) Uniqueness: The answer to the question must be a single, well-defined entity; (b) Complexity: Direct derivation of the entity from the question must require at least one additional web search step;
(3) \textbf{Question Fusion}: Claude-sonnet-4 fuses the deep sub-question with the original wide search query; and (4) \textbf{Human Annotation}: A team of seven master's-level annotators validates and refines the synthesized questions to ensure uniqueness, complexity, and linguistic naturalness. This process requires approximately 40 minutes of human annotation per instance, maintaining the high-quality standards essential for a rigorous benchmark.
The prompts of core entity extraction, deep sub-question synthesis and question fusion are placed at Appendix~\ref{sec:appendix_prompt}.

\subsection{Data Statistics}

Table \ref{tab:benchmark_comparison} provides a comprehensive comparison of our DeepWideSearch benchmark against existing datasets across multiple dimensions. Our benchmark demonstrates significantly higher search complexity compared to prior work, with an average table volume of 414.10 information units, substantially exceeding deep search benchmarks like GAIA and BrowseComp. Crucially, DeepWideSearch requires 4.21 average search steps to identify core entities—nearly 4× more complex than WideSearch (1.24). The dataset spans 15 diverse domains, covering both English and Chinese queries, with 220 carefully curated instances (85 from Deep2Wide, 135 from Wide2Deep). 
These statistics empirically validate the deep and wide attributes of our proposed DeepWideSearch. Cases and more details about the data in Table~\ref{tab:benchmark_comparison} can be found in Appendix~\ref{subsec.data_stats}.


\begin{table*}[htbp]
\centering
\resizebox{\textwidth}{!}{
    \begin{tabular}{ccccccc}
    \toprule
    \textbf{Benchmarks} & 
    \textbf{Domains} & 
    \textbf{Data Size} & 
    \textbf{\begin{tabular}{c}Avg. Sample\\Per Domain\end{tabular}} & 
    \textbf{\begin{tabular}{c}Table\\Volume\end{tabular}} & 
    \textbf{\begin{tabular}{c}Avg. Steps \\Search Entity\end{tabular}}&
    \textbf{Lang.}\\
    \midrule
    TriviaQA~\cite{joshi-etal-2017-triviaqa} & - & 95K & - & 1 & $\approx$ 1 & EN \\
    HotpotQA~\cite{yang2018hotpotqadatasetdiverseexplainable} & - & 113K & - & 1 & $\approx$ 2 & EN \\
    GAIA~\cite{mialon2023gaiabenchmarkgeneralai} & - & 103 & - & 1 & 7.73 & EN \\
    BrowseComp~\cite{wei2025browsecompsimplechallengingbenchmark} & 9 & 1266 & 126.6 & 1 & - & EN \\
    BrowseComp-zh~\cite{zhou2025browsecompzhbenchmarkingwebbrowsing} & 11 & 289 & 26.27 & 1 & - & ZH \\
    WideSearch~\cite{wong2025widesearch} & 14 & 200 & 12.80 & 450.67 & 1.24 & EN,ZH \\ 
    \midrule
    \multicolumn{7}{c}{Our Proposed DeepWideSearch} \\
    \midrule
    Deep2Wide & 15 & 85 & 7.08 & 247.74 & 3.22 & EN,ZH \\
    Wide2Deep & 13 & 135 & 10.38 & 518.84 & 4.55 & EN,ZH \\
    Overall & 15 & 220 & 14.67 & 414.10 & 4.21 & EN,ZH \\
    \bottomrule
    \end{tabular}
}
\caption{Data statistics comparison across benchmarks. GAIA refers to the text-only split.\label{tab:benchmark_comparison}}
\end{table*}

\subsection{Evaluation Metrics of DeepWideSearch\label{subsec.evaluation_metric}}

As shown in Figure~\ref{fig:intro_case_illustration}, we evaluate agent performance on DeepWideSearch along three complementary axes: Depth, Width, and Efficiency.

\paragraph{Depth Evaluation}

The depth dimension evaluate the capability of agents to correctly identify target entities through deep reasoning over multi-hop retrieval. Following previous works~\cite{wei2025browsecompsimplechallengingbenchmark,mialon2023gaiabenchmarkgeneralai}, we introduce the \textbf{Column-F1} metric. As shown in Figure~\ref{fig:intro_case_illustration}, Column-F1 is computed as the F1 score over the unique columns in the table. These unique columns correspond to the core attributes of entities (\textit{i.e.}, rows) that uniquely identify them. Therefore, Column-F1 can be seen as the extension of the accuracy used in established deep search benchmarks, computing the precision of a group of entities (rows in the table).
Higher Column-F1 scores indicate more precise entities identification across the entire table structure.
Moreover, since our proposed two methods include the core entity of questions, we also introduce the \textbf{Core Entity Accuracy (CE Acc.)}, serving as an additional indicator of deep reasoning capability.

\paragraph{Width Evaluation}

The width dimension measures how comprehensively and accurately the agent retrieves all associated information units for entities (rows in the table). Building upon the evaluation framework of WideSearch~\cite{wong2025widesearch}, we assess performance at three granularities: (1) \textbf{Success Rate}: A binary metric indicating whether the agent's output table exactly matches the human-annotated ground truth (all rows, columns, and values identical); (2) \textbf{Row-level F1}: Computes precision, recall, and F1 scores at the row level (i.e., for each entity and its associated attributes), capturing whether the agent retrieves complete contextual information per entity; (3) \textbf{Item-level F1}: The finest-grained metric evaluating accuracy at the individual cell level, reflecting fidelity in retrieving atomic information units within the structured table.

\paragraph{Efficiency Evaluation}
To address the substantial computational costs inherent in web-scale tool usage (including search, browsing APIs), we further evaluate system efficiency through two metrics: (1) \textbf{Input/Output Token}: The total tokens consumed during reasoning and tool calls; (2) \textbf{Cost}: Estimated cost expenditure based on standard model inference API pricing during query resolution. These efficiency metrics are critical for real-world deployment considerations, particularly given the demanding requirements for extensive multi-round search and browsing.

To account for stochasticity in LLM-based agent behavior, we conduct four independent runs per question for each baseline system. For both depth and width metrics, we report three complementary statistics: (1) \textbf{Avg@4}: The mean performance across all four runs; (2) \textbf{Max@4}: The best performance observed across the four runs; and (3) \textbf{Pass@4}: The proportion of questions solved successfully in at least one run (only for Success Rate). This comprehensive evaluation protocol ensures robustness against sampling variance while also highlighting the system's peak performance potential.

\section{Experiments}

\subsection{Experimental Setup}

We evaluate three kinds of baselines on our proposed DeepWideBenchmark: (1) \textit{Closed-source LLMs (without tool calls)}: OpenAI o3-mini, GPT-4o, GPT-5, Claude-sonnet 4, Gemini 2.5 Pro and Qwen-Max; (2) \textit{Open-source LLMs (without tool calls)}: DeepSeek-V3/R1~\cite{guo2025deepseek,liu2024deepseek}, KIMI-K2~\cite{kimiteam2025kimik2openagentic}, Qwen3 series~\cite{yang2025qwen3technicalreport}; and (3) \textit{Open-source Agent Systems}: WebSailor~\cite{li2025websailor},  Smolagents~\cite{smolagents} and OWL~\cite{hu2025owloptimizedworkforcelearning} are equipped with advanced GPT-5, Claude-sonnet-4 and Gemini-2.5-Pro backbone models. All agent systems utilized identical tools: (1) Google Search API; and (2) Webpage Visit tool. Since webpages in HTML format are often very lengthy, we use the same LLM in the agents to summarize the HTML into a concise summarization. The cost of this summarization process is also counted into the efficiency metrics.
We utilized the official API endpoints of these LLMs with their default decoding parameter settings.

\subsection{Main Results}

\begin{table*}[t]
\centering
\resizebox{1\linewidth}{!}{
\begin{tabular}{l|cc cc cc | cc cc}
\toprule
\textbf{Model / System} & \multicolumn{2}{c}{\textbf{Success Rate (\%)}} & \multicolumn{2}{c}{\textbf{Row F1 Score (\%)}} & \multicolumn{2}{c|}{\textbf{Item F1 Score (\%)}} & \multicolumn{2}{c}{\textbf{Column F1 (\%)}} & \multicolumn{2}{c}{\textbf{CE Acc. (\%)}}\\
\cmidrule(lr){2-3} \cmidrule(lr){4-5} \cmidrule(lr){6-7} \cmidrule(lr){8-9} \cmidrule(lr){10-11}
& Avg@4 & Pass@4 & Avg@4 & Max@4 & Avg@4 & Max@4 & Avg@4 & Max@4 & Avg@4 & Pass@4 \\
\midrule
\multicolumn{11}{c}{\textit{Closed-source LLMs}} \\
\midrule
OpenAI o3-mini &0.0&0.0&3.35&4.55&13.59&16.85&27.36&35.68&61.59&69.55\\
GPT-5 &0.30&1.36&9.61&13.42&21.67&28.21&31.71&41.05&58.41&72.72\\
Claude Sonnet 4 &0.9&0.9&7.31&8.97&19.94&23.38&32.63&40.16&57.95&64.09\\
Gemini 2.5 Pro &0.9&1.82&15.42&18.96&32.06&37.10&\textbf{45.27}&52.86&\textbf{73.98}&81.82\\
Qwen-Max &0.0&0.0&4.16&6.18&14.32&18.48&28.81&36.19&56.02&63.64\\
GPT-4o & 0.0 & 0.0 & 4.18 &7.01 & 11.86 & 16.41 &19.66&27.07&54.20&63.64\\
\midrule
\multicolumn{11}{c}{\textit{Open-source LLMs}} \\
\midrule
DeepSeek-V3 & 0.23 & 0.45 & 6.52 & 9.99 & 19.08 &24.32 & 31.26&39.56& 60.68 & 69.09 \\
DeepSeek-R1 &0.28&0.45&10.72&14.39&25.01&30.56&38.42&47.77&66.93&75.45\\
KIMI-K2 &0.34&0.91&7.74&11.92&20.44&27.54&31.48&41.83&64.32&73.18\\
Qwen3-235B-A22B &0.0&0.0&2.94&5.74&12.38&19.53&22.03&34.99&52.39&67.73\\
Qwen3-235B-A22B-Instruct &0.0 &0.0 &3.50 &5.34 & 13.28 & 17.85 &24.64&33.03&56.82 &64.09 \\
Qwen3-32B &0.0&0.0&2.28&3.67&12.05&16.26&26.37&35.97&54.66&66.36\\
\midrule
\multicolumn{11}{c}{\textit{Open-source Agent Framework with Advanced LLMs}} \\
\midrule
OWL (Gemini 2.5 Pro) &0.0&0.0&11.11&16.93&28.75&41.70&34.84&50.39&66.14&81.82\\
OWL (Claude sonnet 4) &0.68&1.36&8.29&14.81&20.44&31.65&30.08&45.50&67.39&81.82\\
Smolagents (Gemini 2.5 Pro) &0.11&0.45&9.01&15.65&18.53&30.91&27.39&45.09&60.00&79.09\\
Smolagents (Claude sonnet 4) &0.91&0.91&5.06&8.94&14.49&22.68&21.60&33.83&62.95&74.09\\
Smolagents (GPT-5) &0.45&0.45&8.18&14.27&20.26&30.66&31.83&44.41&66.48&80.00\\
WebSailor (Gemini 2.5 Pro) &1.25&2.73&12.51&20.49&25.29&39.11&34.41&52.69&70.57&81.36\\
WebSailor (Claude Sonnet 4) &\textbf{2.39}&\textbf{3.64}&\textbf{16.88}&\textbf{24.26}&\textbf{32.90}&\textbf{42.35}&42.01&\textbf{54.01}&70.91&80.90\\
WebSailor (GPT-5) &0.34&1.36&10.97&16.17&25.96&35.65&37.18&49.48&74.32&\textbf{85.00}\\
\bottomrule
\end{tabular}
}
\caption{Main results on our proposed DeepWideSearch benchmark.}
\label{tab:main_results}
\end{table*}

The complete results are presented in Table~\ref{tab:main_results}. It can be found that most baselines demonstrate near-zero success rates, with only WebSailor (Gemini 2.5 Pro) and WebSailor (Claude Sonnet 4) exceeding 1-2\% in Success Rate (Avg@4), confirming the inherent complexity of simultaneously handling deep reasoning and wide-scale information collection. Notably, Gemini 2.5 Pro emerges as the top-performing LLM, achieving the highest Column F1 (45.27\%, Avg@4), Core Entity Accuracy (73.98\%, Avg@4), and Pass@4 Success Rate (1.82\%), even outperforming several agent systems. This exceptional performance indicates that Gemini 2.5 Pro possesses advanced reasoning capabilities for entity identification and extensive internal knowledge for filling result tables without external search.
Furthermore, we detail the performance of baselines from depth and width metrics as below.

\paragraph{Depth Metrics}
Our analysis reveals that agent systems generally enhance the deep search capabilities of base LLMs, as evidenced by consistent improvements in Core Entity Accuracy (CE Acc.). For example, the CE Acc. (Avg@4) of GPT-5 increases from 58.41\% (base LLM) to 74.32\% when integrated into WebSailor, representing a +15.91 percentage point gain. Similarly, Claude Sonnet 4 improves from 57.95\% to 70.91\% under WebSailor, demonstrating the effectiveness of iterative tool calls and multi-step reasoning in complex information retrieval. However, Gemini 2.5 Pro represents a notable exception to this trend. Upon close inspection of generated outputs, we find that Gemini 2.5 Pro in agent systems frequently fails due to three critical issues: (a) producing invalid markdown-formatted tables; (b) executing incorrect tool call APIs; and (c) incomplete task solving due to inference errors, occurring in 24.24\% of cases on average—substantially higher than GPT-5 (16.36\%) and Claude Sonnet 4 (17.80\%). This suggests that Gemini 2.5 Pro's output formatting behavior becomes brittle when subjected to multi-step tool orchestration.
Critically, while agent systems improve core entity identification, they fail to consistently enhance column-level precision. For instance, the Column F1 (Avg@4) of  Claude Sonnet 4 model declines from 32.63\% (base LLM) to 30.08\% in OWL and 21.60\% in Smolagents. This pattern highlights a fundamental limitation: even when agents successfully identify core entities through multi-hop reasoning, current agent architectures cannot reliably collect complete entities, with their effectiveness often falling below the usage of internal knowledge in base LLMs.


\paragraph{Width Metrics}


When evaluating width metrics that measure comprehensive information collection, we observe that most agent frameworks do not significantly improve the base LLMs' wide search capabilities. Only three combinations demonstrate consistent improvements across all width metrics: OWL (Claude Sonnet 4), WebSailor (Claude Sonnet 4), and WebSailor (GPT-5). The remaining agents show substantial performance degradation compared to their counterpart base LLMs.
Beyond the issues specific to Gemini 2.5 Pro that described above, the Smolagents framework also consistently underperforms across nearly all metrics. 
Our investigation reveals that Smolagents employs minimal reasoning before tool calls, which restricts the effectiveness of subsequent tool calls. This architectural constraint prevents Smolagents from formulating precise search queries, resulting in inadequate information coverage and poor performance on width metrics.

\section{Analysis}

In this section, we conduct several detailed analysis on Efficiency (Section~\ref{subsec.efficiency_analysis}), Tool Calls (Section~\ref{subsec.tool_call_analysis}), Differences in Dataset Construction Methods (Section~\ref{subsec.difference_in_dataset_method}), Per-topic Performance (Section~\ref{subsec.per_topic_analysis}), and Error Analysis (Section~\ref{subsec.error_analysis}).

\subsection{Efficiency Analysis\label{subsec.efficiency_analysis}}

\begin{wraptable}[11]{r}{0.5\textwidth}
\vspace{-12pt}
\centering
\caption{Average token usage and cost statistics for some agents on DeepWideSearch questions.}
\label{tab:token_cost}
\resizebox{1\linewidth}{!}{
\begin{tabular}{l r r r}
\toprule
\textbf{Agents} & \textbf{\begin{tabular}{@{}c@{}}Input\\Token\end{tabular}} & \textbf{\begin{tabular}{@{}c@{}}Output\\Token\end{tabular}} & \textbf{Cost (\$)} \\
\midrule
OWL (Gemini 2.5 Pro)          & 65K   & 2.5K  & $\approx 0.2$ \\
OWL (GPT-5)                   & 1.8M  & 50K   & $\boldsymbol{\approx 2.75}$ \\
Smolagents (Claude Sonnet 4)  & 224K  & 2.4K  & $\approx 2.14$ \\
Smolagents (GPT-5)            & 120K  & 25K   & $\approx 0.90$ \\
WebSailor (Gemini 2.5 Pro)    & 65K   & 2.5K  & $\approx 0.49$ \\
WebSailor (Claude Sonnet 4)   & 186.2K& 3.5K  & $\approx 1.40$ \\
WebSailor (GPT-5)             & 17.7K & 6.2K  & $\approx 0.36$ \\
\bottomrule
\end{tabular}
}
\end{wraptable}

Compared to deep search or wide search, DeepWideSearch imposes significantly higher computational and operational overhead. 
As shown in Table~\ref{tab:token_cost}, even state-of-the-art agents incur substantial resource costs per query. For instance, OWL (GPT-5) and WebSailor (Claude Sonnet 4) achieve average \$2.75 and \$1.40 per question — with many queries remaining unresolved despite this high cost.
Due to unstable network conditions and tool call errors, agents often require multiple retry attempts to complete tasks such as search, significantly increasing computational overhead—for instance, OWL (GPT-5) incurs an average cost exceeding \$6.8 under retry conditions.
These results underscore a critical inefficiency in current agent architectures when tackling complex deep and wide information seeking tasks. This suggests that existing systems are not yet scalable for real-world deployment of DeepWideSearch, motivating future work on efficient planning, memory reuse, and adaptive resource allocation.

\subsection{Tool Calls Analysis\label{subsec.tool_call_analysis}}


\begin{wraptable}[7]{r}{0.4\textwidth}
\vspace{-12pt}
\centering
\caption{Average tool calls in the WebSailor system.}
\label{tab:tool_call}
\resizebox{\linewidth}{!}{
\begin{tabular}{l r r}
\toprule
\textbf{Agents} & \textbf{Search} & \textbf{Visit} \\
\midrule
WebSailor (Gemini 2.5 Pro)    & 4.77 & 2.68 \\
WebSailor (Claude Sonnet 4)   & 23.23 & 4.57 \\
WebSailor (GPT-5)             & 8.72 & 5.35 \\
\bottomrule
\end{tabular}
}
\end{wraptable}

Table~\ref{tab:tool_call} shows the average number of tool calls (Search and Visit tools) per sample across different backbone LLMs in WebSailor. Notably, WebSailor (Claude Sonnet 4) exhibits a significantly higher Search tool calls (23.23) compared to Gemini 2.5 Pro (4.77) and GPT-5 (8.72). This aligns with its superior performance (Table~\ref{tab:main_results}), suggesting that scaling the search tool calls improves the performance.

\begin{table}[t]
\centering
\resizebox{1\linewidth}{!}{
\begin{tabular}{l|cc cc cc|cc cc}
\toprule
\textbf{Model / System} & \multicolumn{2}{c}{\textbf{Success Rate (\%)}} & \multicolumn{2}{c}{\textbf{Row F1 Score (\%)}} & \multicolumn{2}{c}{\textbf{Item F1 Score (\%)}} & \multicolumn{2}{|c}{\textbf{Column F1 (\%)}} & \multicolumn{2}{c}{\textbf{Entity Acc. (\%)}}\\
\cmidrule(lr){2-3} \cmidrule(lr){4-5} \cmidrule(lr){6-7} \cmidrule(lr){8-9} \cmidrule(lr){10-11}
& Avg@4 & Pass@4 & Avg@4 & Max@4 & Avg@4 & Max@4 & Avg@4 & Max@4 & Avg@4 & Pass@4 \\
\midrule
\multicolumn{11}{c}{\textit{Wide Search $\boldsymbol{\rightarrow}$ DeepWideSearch} (Wide2Deep)} \\
\midrule
Avg. LLMs & 1.17 & 2.22 & 17.23 & 21.80 & 38.04 & 43.95 & 50.94 & 59.09 & 90.12 & 93.83 \\
Avg. Agents & 1.23 & 2.13 & 15.55 & 24.13 & 33.51 & 46.98 & 44.13 & 60.70 & 88.36 & 96.76\\
Avg. All & 1.21 & 2.15 & 16.00 & 23.49 & 34.75 & 46.16 & 45.96 & 60.26 & 88.84 & 95.96 \\
\midrule
\multicolumn{11}{c}{\textit{Deep Search $\boldsymbol{\rightarrow}$ DeepWideSearch} (Deep2Wide)} \\
\midrule
Avg. LLMs & 0.0 & 0.0 & 2.67 & 3.92 & 8.52 & 13.25 & 13.67 & 21.81 & 31.77 & 46.27 \\
Avg. Agents & 0.15 & 0.44 & 3.25 & 5.99 & 9.21 & 16.43 & 13.75 & 24.92 & 33.86 & 54.56 \\
Avg. All & 0.11 & 0.32 & 3.09 & 5.42 & 9.02 & 15.56 & 13.73 & 24.07 & 33.29 & 52.30 \\
\midrule
\multicolumn{11}{c}{Overall} \\
\midrule
Avg. LLMs & 0.72 & 1.36 & 11.60 & 14.89 & 26.64 & 32.09 & 36.54 & 44.69 & 67.58 & 75.45 \\
Avg. Agents & 0.75 & 1.25 & 9.36 & 14.88 & 20.76 & 30.60 & 27.77 & 40.74 & 58.05 & 69.89 \\
Avg. All & 0.74 & 1.28 & 9.97 & 14.88 & 22.36 & 31.01 & 30.16 & 41.82 & 60.65 & 71.40 \\
\bottomrule
\end{tabular}
}
\caption{Performance comparison between Deep2Wide and Wide2Deep methods.\label{tab:diffiulty_of_data_splits}}
\end{table}


\subsection{Differences in Dataset Construction Methods\label{subsec.difference_in_dataset_method}}
Table~\ref{tab:diffiulty_of_data_splits} demonstrates the  average performance of advanced LLMs (GPT-5, Claude Sonnet 4 and Gemini 2.5 Pro) with their counterpart agent systems.
It can be found that the Deep2Wide construction method produces substantially more challenging data than Wide2Deep method. For example, LLMs and agents achieves nearly 0.0\% success rate on Deep2Wide (Avg. LLMs: 0.0\% Avg@4; Avg. Agents: 0.15\% Avg@4), compared to the Wide2Deep (Avg. LLMs: 1.17\% Avg@4; Avg. Agents: 1.23\% Avg@4). Critically, the overall Entity Accuracy on Deep2Wide is only 33.29\% (vs. 88.84\% on Wide2Deep). This observation indicates that the synthesized deep sub-question in the Wide2Deep method is easier for LLMs to solve.
Nevertheless, the column-F1 of Wide2Deep remains below 51\%, indicating that comprehensively collecting entities is still challenging.

\subsection{Per-topic Performance Analysis\label{subsec.per_topic_analysis}}


\begin{figure*}[ht]
    \centering
    \includegraphics[width=\linewidth]{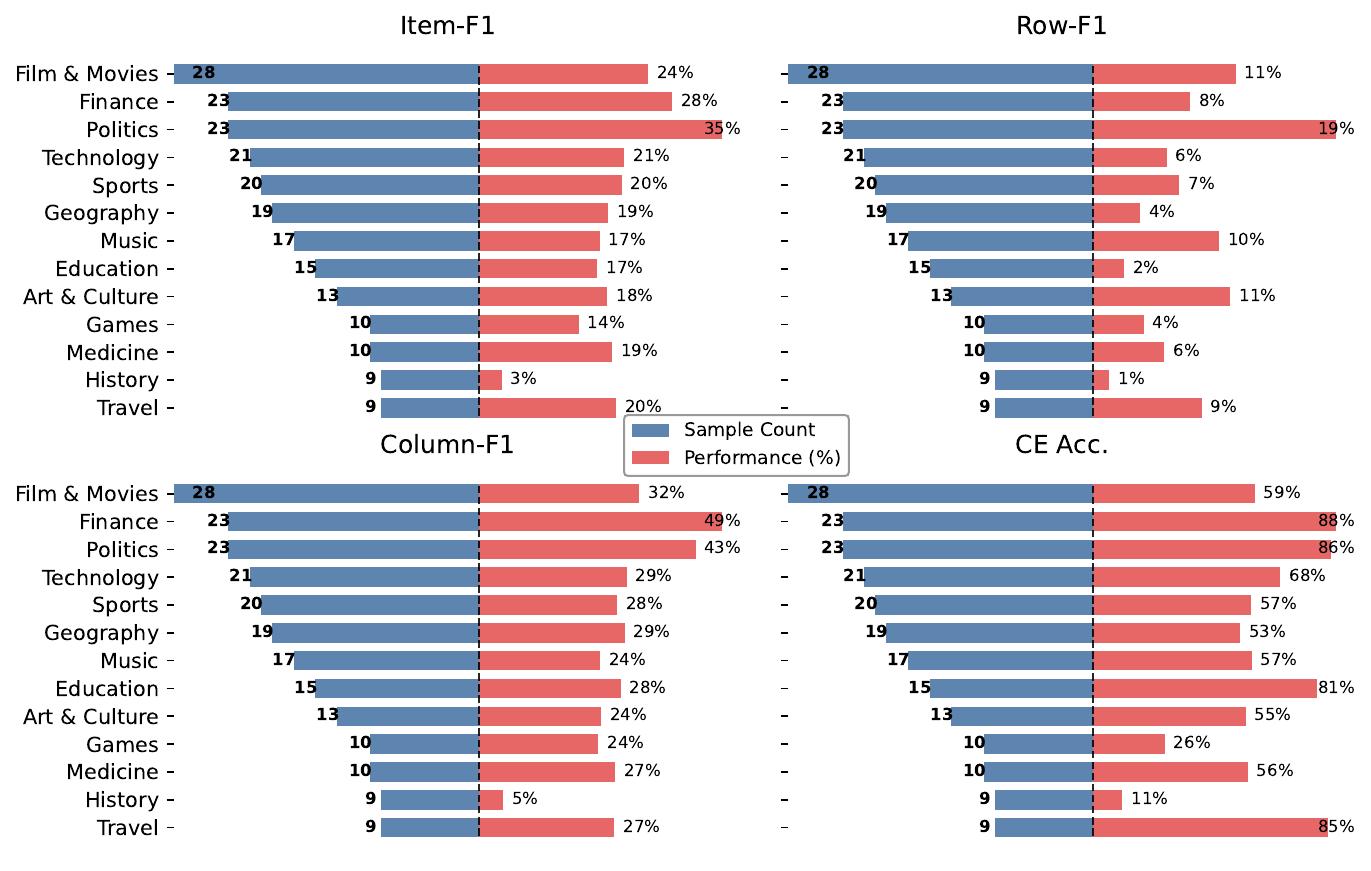}
    \caption{Per-topic analysis on two depth metrics (Column F1 and CE Acc.) and two width metrics (Item F1 and Row F1).}
    \label{fig:per_topic_analysis}
\end{figure*}


As shown in Figure~\ref{fig:per_topic_analysis}, we analyze topic-wise performance through bidirectional bar charts evaluating depth metrics (Column-F1, CE Acc.) and width metrics (Item-F1, Row-F1), excluding domains with fewer than 5 samples. 
Four key patterns emerge: (1) The top-5 most frequent topics (sample count >20) are \textit{Film \& Movies}, \textit{Politics}, \textit{Finance}, \textit{Technology}, and \textit{Sports};
(2) \textit{Politics} achieves the highest item- and row-level F1 scores (35\% and 19\%), indicating wide search are more tractable in this topic, while \textit{Politics} and \textit{Finance} attain the highest column F1 and CE accuracy, suggesting deep search are comparatively easier here;
(3) Despite strong depth performance in \textit{Finance}, \textit{Travel}, and \textit{Education} topics, the performance of baselines exhibit substantially lower width metrics on these three topics (e.g., \textit{Travel} 20\% item F1 and \textit{Finance} 8\% row F1), revealing that strong deep search capability does not guarantee effective wide search capability;
and (4) \textit{History} and \textit{Games} consistently underperform across all metrics (\textit{e.g.}, 5\% Column-F1 of \textit{History}), establishing them as the most challenging topics. These findings highlight the heterogeneous nature of search complexity across topics.

\subsection{Error Analysis\label{subsec.error_analysis}}

As shown in Tables~\ref{tab:main_results}, agent systems might underperform backbone LLMs on DeepWideSearch tasks. Our error analysis reveals four key failure patterns:
(1) \textbf{Lack of Reflection:} agents often lack effective reflection mechanisms. When encountering wrong trajectories (Figures~\ref{fig:case_lack_of_reflection_1}) or tool call errors (Figure~\ref{fig:case_lack_of_reflection_2}), they prematurely conclude the task is unsolvable and output empty tables rather than analyzing failure causes and exploring alternative paths;
(2) \textbf{Overreliance on Internal Knowledge:} agents frequently overrely on internal knowledge. Even when correctly identifying core entities (Figure~\ref{fig:case_internel_knowledge}), they often generate tables solely using their internal parametric knowledge rather than performing proper web queries, resulting in outdated or inaccurate information due to limited training data scope;
(3) \textbf{Insufficient Retrieval:} information retrieval is often insufficient. For example, despite identifying relevant pages (Figure~\ref{fig:case_insufficient_retrieval}), agents frequently fail to properly access complete context through visit operations, leading to significant information omissions. Even when visit operations are executed correctly, summarized webpage data may still miss critical details. This limitation motivates the design of a question-aware, customized webpage summarization process in agent systems;
and (4) \textbf{Context Overflow:} context overflow presents a fundamental challenge. Deep wide search requires extensive multi-step reasoning and numerous search tool calls, significantly expanding context length (Figure~\ref{fig:case_context_overflow}). This issue occurred in 24.96\% of cases, exceeding the context management capabilities of current agent architectures;
In summary, these four error patterns highlight that current agents face substantial limitations when addressing the challenges of depth and width in complex information-seeking tasks. Addressing these limitations requires specialized architecture for deep wide search scenarios.



\section{Conclusion}

This paper addresses the critical gap in information-seeking agent evaluation by introducing DeepWideSearch benchmark, the first benchmark designed to simultaneously assess deep reasoning and wide-scale information collection. Our experiments demonstrate that state-of-the-art agents achieve only 2.39\% average success rate on this challenging benchmark, revealing fundamental limitations for current agents. These results underscore the combinatorial complexity of deep and wide search as a key frontier to guide future research toward more capable information-seeking agents.

\section{Limitations and Future Work}
Despite our established DeepWideSearch benchmark, there are three key limitations remain to be addressed in the future work:
(1) As shown in Table~\ref{tab:diffiulty_of_data_splits}, the Wide2Deep construction method produces significantly easier questions than Deep2Wide, as evidenced by the substantially higher CE Accuracy. We will iteratively refine sub-questions to increase question complexity while maintaining natural language quality;
(2) Our current dataset exhibits slight differences with real-world deep and wide search questions in terms of solution paths (Cases in Appendix~\ref{subsec.limitation}). In future work, we will iteratively refine the DeepWideSearch dataset to better align with real-world applications;
and (3) Our dataset construction relies heavily on human annotation, limiting scalability. Future work should explore automated data generation techniques and develop reference-free evaluation metrics that avoid complex, human-verified tabular answers, enabling efficient dataset expansion and model optimization across diverse domains.

\bibliography{references,custom}

\appendix 

\clearpage
\newpage
\section{Details of Datasets\label{subsec.data_stats}}
The table volume in Table~\ref{tab:benchmark_comparison} represents the number of the searched information in the DeepWideSearch questions, which is defined as the product of rows and columns of the table.
The average steps of the search entities is counted as the number of the reasoning steps and tool calls. Specifically, the average steps of GAIA is counted by the reference trajectories in the dataset, and the average steps of WideSearch is annotated by our three human raters.
Besides, Figure~\ref{fig:one_case_in_deepwidesearch} and Figure~\ref{fig:two_case_in_deepwidesearch} present two cases in our proposed DeepWideSearch dataset.

\begin{figure*}[htbp]
    \begin{tcolorbox}[
        title=Case Deep2Wide\_62 Instance (Core Entity is Nobel Proze),
        colback=yellow!10,
        colbacktitle=black!80,
        coltitle=white,
        arc=1mm,
        boxrule=0pt,
        breakable]
        \textbf{DeepWide Question}: Soviet physicist A received his Ph.D. at the age of 27 under the supervision of the renowned Soviet physicist B. Physicist B was awarded the Nobel Prize at the age of 54 and passed away six years later. In 2023, physicist A received a prestigious international prize in physics. Please provide the names, specific award titles, dates of birth (formatted as "Mon DD, yyyy"; if the exact date is unknown, use "-, yyyy"), and nationalities of the other scientists who received awards in the same year as A. Present the results in a single Markdown table with the following columns in order: Name, Award, Date of Birth, Nationality. All cells must be filled according to the column requirements; do not omit any information arbitrarily. The table must be output entirely in Chinese, and the final output should strictly follow the format: \verb|```|markdown{table content}\verb|```|.
    \end{tcolorbox}
    
    \begin{tcolorbox}
    [
        title=Ground-truth Table of Deep2Wide\_62 Instance,
        colback=green!10,
        colbacktitle=black!80,
        coltitle=white,
        arc=1mm,
        boxrule=0pt,
        breakable]
    \begin{center}
    \small
        \begin{tabular}{|c|c|c|c|}
        \toprule
        \textbf{Name} & \textbf{Award} & \textbf{Date of Birth} & \textbf{Nationality} \\\midrule
Peter Agre & Chem & Jan 30, 1949 & United States \\
Roderick MacKinnon & Chem & Feb 19, 1956 & United States \\
Clive Granger & Econ & Sep 04, 1934 & United Kingdom \\
Robert Engle & Econ & Nov 10, 1942 & United States \\
Alexei Abrikosov & Phys & Jun 25, 1928 & Russia/United States \\
Vitaly Ginzburg & Phys & Oct 04, 1916 & Russia/United States \\
Anthony Leggett & Phys & Mar 26, 1938 & United Kingdom/United States \\
J. M. Coetzee & Lit & Feb 09, 1940 & South Africa \\
Shirin Ebadi & Peace & Jun 21, 1947 & Iran \\
Paul Lauterbur & Med & May 06, 1929 & United States \\
Peter Mansfield & Med & Oct 09, 1933 & United Kingdom \\
        \bottomrule
        \end{tabular}
        \end{center}
    \end{tcolorbox}

\caption{One case in DeepWideSearch dataset.}
\label{fig:one_case_in_deepwidesearch}
\end{figure*}

\clearpage
\newpage

\begin{figure*}[htbp]
    \begin{tcolorbox}[
        title=Case Deep2Wide\_26 Instance (Core Entity is Chen Yixun),
        colback=yellow!10,
        colbacktitle=black!80,
        coltitle=white,
        arc=1mm,
        boxrule=0pt,
        breakable]
        \textbf{DeepWide Question}: A pop song performed by a well-known Chinese singer, in which the song title appears ten times in the lyrics, and the first two words of the album title start with the same letter. The lyricist once served as a judge on a music variety show and later appeared on a local TV program alongside another famous lyricist. Could you please list the TV dramas the singer has acted in, and summarize their titles, directors, and main cast? Present the results in a Markdown table with the columns in the following order: Name, Director, Main Cast. Output only the result in the format: \verb|```|markdown{table content}\verb|```|.
    \end{tcolorbox}
    
    \begin{tcolorbox}
    [
        title=Ground-truth Table of Deep2Wide\_26 Instance,
        colback=green!10,
        colbacktitle=black!80,
        coltitle=white,
        arc=1mm,
        boxrule=0pt,
        breakable]
    \begin{center}
    \small
        \begin{tabular}{|c|c|c|}
        \toprule 
    \textbf{Name} & \textbf{Director} & \textbf{Main Cast} \\\midrule
    Brief Marriage & Chen Zhifa & Eason Chan, Cecilia Yip \\
    Toward Happiness & Lin Jianzhong & Nicky Wu, Tang Yuhong \\
    Bloody Marriage & Jin Ge & Gu Shaohua \\
    Sapphire Night Sky & Lai Hoi Shan & Charlene Choi, Eason Chan \\
    Let's Cheer Together & Zhu Ruibin & Steven Cheung, Kenny Kwan \\
    Triumph in the Skies & Poon Ka Tak & Anthony Wong, Myolie Wu \\
    \begin{tabular}{@{}c@{}}When Four-Leaf Clover \\Meets the Sword Tip\end{tabular} & Zhang Yijie & Lai Lok Yi, Li Rilang \\
    Pet Love & Cheng Kei Sing & Louis Koo, Sammi Cheng \\
    Midnight Express & \begin{tabular}{@{}c@{}}Ono Tetsujiro\\Takemura Kentaro\end{tabular} & Takao Osawa, Nanako Matsushima \\
        \bottomrule
    \end{tabular}
    \end{center}
\end{tcolorbox}
\caption{One case in DeepWideSearch dataset.}
\label{fig:two_case_in_deepwidesearch}
\end{figure*}

\clearpage
\newpage

\section{Differences between Our Dataset and Real-world Questions\label{subsec.limitation}}


\begin{figure*}[htbp]
    \begin{tcolorbox}[
        title=Two Deep and Wide Questions,
        colback=yellow!10,
        colbacktitle=black!80,
        coltitle=white,
        arc=1mm,
        boxrule=0pt,
        breakable]
        \textbf{Our DeepWideSearch Question}: Identify an artist who studied at both the Central Academy of Fine Arts in China and the Kunstakademie Düsseldorf in Germany, and who pursued further studies in Germany. During his time in Germany, he studied under three renowned artists, one of whom held the record for the highest auction price ever achieved by a living artist in 2012. Gather information about this artist’s solo exhibitions, including exhibition title, venue, city, and exhibition dates. \\
        \textbf{Solution Path}: first identify the artist (\textcolor{red}{deep/wide search}), then perform wide search for information collection (\textcolor{red}{wide search}).

        \medskip

        \textbf{Real-world Deep and Wide Question}: Please help me identify emerging local merchants from Thailand and Vietnam that operate pet supplies categories on e-commerce platforms such as Lazada and Shopee, and whose store GMV growth rate exceeded 50\% in the first half of 2025. (Emerging merchants are defined as those that opened their stores after 2023.) I need the following information for each qualifying store: store URL, company legal entity details, business contact email, and the founder’s LinkedIn profile.\\
        \textbf{Solution Path}: first perform wide search to list candidate stores (\textcolor{red}{wide search}), then perform deep search to collect their information (GMV) for verification (\textcolor{red}{deep search}).
    \end{tcolorbox}
    \caption{Two cases of the deep and wide search questions.    \label{fig_deepwidesearch_comparison}}
\end{figure*}

Figure~\ref{fig_deepwidesearch_comparison} illustrates two representative deep and wide search questions: the first is an example from our constructed DeepWideSearch dataset, and the second is drawn from a real-world e-commerce scenario. While our dataset captures the essential characteristics of deep and wide search, the primary difference from real-world settings lies in the solution path. In our dataset, the process emphasizes first performing a deep search to gather critical information, followed by a wide search to expand relevant attributes. In contrast, real-world tasks often begin with a wide search to collect a large pool of candidates, followed by a deep search over each candidate for verification.
Nevertheless, it is important to emphasize that despite this procedural difference, our dataset still exhibits the traits of deep and wide search. Specifically, during the initial deep search phase, the model also need to list and reason over a set of candidates, systematically applying deep verification to determine which candidates satisfy the problem constraints and thereby identify the correct target entity. Consequently, even this first-stage deep search inherently incorporates the characteristic of the wide search.

\section{Prompts for DeepWideSearch Data Construction\label{sec:appendix_prompt}}

This section presents three prompts for Wide2Deep method: (1) Core Entity Extraction Prompt in Figure~\ref{img:entity_extraction_prompt}; (2) Deep Sub-Question Synthesis Prompt in Figure~\ref{img:deep_sub_question_systhesis}; (3) Question Fusion in Figure~\ref{img:question_refinement}.

\begin{figure*}[htbp]
\small
\centering
\begin{tcolorbox}

\textbf{\# You are a professional entity extraction expert tasked with identifying the most critical entity from a given text query.}\\
\\
\textbf{\# Classification Requirements:}\\
- The entity must be a specific, concrete entity object mentioned within the text query.\\
- A query may contain multiple entities, but prioritize outputting the single most central entity. If no single core entity can be determined, output multiple entities.\\
- Output ONLY JSON-formatted data containing the entities, with no additional content, explanations, or numbering (to ensure direct parseability). Please directly output the entity, without any explanation.
\\
\textbf{\# Reference Few-shot Examples}
\\
* Input:\\
I absolutely love Jay Chou. Please find all songs released by Jay Chou between January 2004 and September 2010 (including January 2004 and September 2010). Include song details: title, lyricist, composer, release date, album, and duration.\\Notes:\\1. I want only Jay Chou's original vocals (collaborations allowed), excluding instrumental tracks.\\2. Format release dates as yyyy/mm/dd; duration as x minutes x seconds (e.g., 3 minutes 5 seconds).\\3. Include only songs released in China.\\4. Exclude live versions and demos.\\5. Include only songs from Jay Chou's studio albums (exclude singles, single albums, film/TV soundtracks, EPs).\\
* Output: Jay Chou\\
\\
* Input:\\
I want to register for the 2026 postgraduate entrance exam. Please check the (retest) for the Journalism and Communication program (full-time professional master's) at Chinese universities in Region A with 211 Project status or higher for 2025 (total score only).\\
* Output: Journalism and Communication program\\
\\
* Input:\\
Using statistics from the Stockholm International Peace Research Institute (SIPRI), list the specific military expenditures (in billion USD without decimals, e.g., 9000 billion USD), GDP (in trillion USD to two decimal places, e.g., 30.21 trillion USD), global military expenditure ranking, head of state (actual leader), and defense minister for the US, Russia, Germany, India, and Japan for each year from 2019 to 2024 (inclusive).\\
* Output: Stockholm\\
\\
* Input:\\
I'm running out of books to read. Could you compile a ranked list of the top 10 books from Douban Reading's annual (overall) for 2022-2024 (inclusive), plus the top 10 bestsellers and highest-rated books from Dangdang.com each year? Include authors' names.\\
* Output: Douban Reading, Dangdang.com\\
\\
* Input:\\
Please compile a table listing the "CNN Hero of the Year" for every year the award was actually presented, from its first introduction through 2024 (including 2024), along with relevant details for each honoree.\\
* Output: CNN Hero of the Year\\
\\
\textbf{\# Perform entity extraction for the following query:}\\
\{question\}
\end{tcolorbox}
\caption{The prompt of core entitiy extraction in Wide2Deep method.}
\label{img:entity_extraction_prompt}
\end{figure*}

\begin{figure*}[htbp]
\small
\centering
\begin{tcolorbox}
Please help me gather all available information about ``\{entity\}'', and based on the collected information, synthesize a multi-hop query that meets the following requirements:
\\
(1) The answer to the query must be exactly ``\{entity\}'' only—no other answers or ambiguities allowed;  \\
(2) The information about ``\{entity\}'' included in the query should not be overly specific (e.g., exact dates, locations, awards, or distinctive features), so that one cannot directly find ``{entity}'' by searching those fragments online; \\ 
(3) A human answering the query must perform multiple search steps and gather information from at least three distinct, non-repeating URLs to logically infer ``\{entity\}'';\\
(4) The query must be concise. Avoid constructing long queries by piling up excessive features. Include only 2–3 pieces of information about \{entity\}. Focus on the ambiguity and reliability of these features, rather than the quantity; \\
(5) Consider temporal validity—the answer to the query must remain stable over time and not change with time; \\
(6) After generating the query, you must conduct an additional verification process via search engines: extract 3–5 simplified search queries from the synthesized query (each reflecting one key feature or phrase from the original query; keep them short and focused on core keywords). Analyze the search results to ensure that ``\{entity\}'' cannot be directly found in a single step. \textbf{If ``\{entity\}'' can be found directly through any of these 3–5 searches, the synthesized query is too simple and does not meet requirements—please repeat the entire process until none of the derived searches can directly reveal ``\{entity\}''.}\\
\\
\textbf{\# Output Format: Place all generated query, reference URLs, and reasoning in the standard JSON structure}\\
\end{tcolorbox}
\caption{The prompt of deep sub-question synthesis in Wide2Deep method.}
\label{img:deep_sub_question_systhesis}
\end{figure*}

\begin{figure*}[htbp]
\small
\centering
\begin{tcolorbox}
You are a senior question synthesis expert responsible for integrating a complex question about an entity into an original query containing that entity, thereby constructing a more challenging composite query. Please strictly follow the rules below:
\\
1. \textbf{Task Objective}:\\
   - Replace the entity in the original query with the corresponding complex question.\\
   - The generated query must require users to first resolve the complex question to obtain the entity information, then proceed with the information retrieval steps in the original query.\\
   - The final synthesized query must be grammatically correct and logically coherent.\\
   - The complex question is in Chinese, but the original query may be in either Chinese or English: if the original query is in English, you must translate the complex question into accurate and equivalent English and integrate it into the query (without adding or omitting any information from the original Chinese question, ensuring full translation fidelity); if the original query is in Chinese, the final synthesized query must also be in Chinese.\\
\\
2. \textbf{Input Specifications}:\\
   - Original query (query): \{query\}\\
   - Entity (entities): {entities} (for internal reference only; \textbf{must not appear in the output})\\
   - Complex question (question): A complex query question to find {entities}: {question}\\
\\
3. \textbf{Synthesis Rules}:\\
   - Replace the entity in the original query with the descriptive text of the complex question.\\
   - The complex question must be transformed into a noun phrase (remove the question mark and rephrase it as a descriptive clause).\\
   - \textbf{Do not} reveal any entity information (specific names from ``entities'' must not appear).\\
   - Maintain professional translation quality and avoid awkward or unnatural phrasing.\\
\\
\textbf{Output only the new synthesized query that integrates the complex question with the original query, without any explanations or additional content}
\end{tcolorbox}
\caption{The prompt of deep and wide question fusion in Wide2Deep method.}
\label{img:question_refinement}
\end{figure*}

\newpage

\section{Error Cases in DeepWideSearch}
\label{sec:case}


This section provides the four kinds of representative errors of agents: (1) Lack of Reflection (Figure~\ref{fig:case_lack_of_reflection_1} and Figure~\ref{fig:case_lack_of_reflection_2}); (2) Overreliance on Internal Knowledge (Figure~\ref{fig:case_internel_knowledge}); (3) Context Overflow (Figure~\ref{fig:case_context_overflow}); and (4) Insufficient Retrieval (Figure~\ref{fig:case_insufficient_retrieval}).

\begin{figure*}[htbp]
\begin{center}\tcbox[colback=gray!20,colframe=gray!60,boxrule=0.6pt,arc=1mm]{\bfseries Error: Lack of Reflection}\end{center}\vspace{2mm}
    \begin{tcolorbox}[
        title=Input of Deep2Wide\_62 Instance,
        colback=yellow!10,
        colbacktitle=black!80,
        coltitle=white,
        arc=1mm,
        boxrule=0pt,
        breakable]
        \textbf{DeepWide Question}: Soviet physicist A received his Ph.D. at the age of 27 under the supervision of the renowned Soviet physicist B. Physicist B was awarded the Nobel Prize at the age of 54 and passed away six years later. In 2023, physicist A received a prestigious international prize in physics. Please provide the names, specific award titles, dates of birth (formatted as "Mon DD, yyyy"; if the exact date is unknown, use "-, yyyy"), and nationalities of the other scientists who received awards in the same year as A. Present the results in a single Markdown table with the following columns in order: Name, Award, Date of Birth, Nationality. All cells must be filled according to the column requirements; do not omit any information arbitrarily. The table must be output entirely in Chinese, and the final output should strictly follow the format: ```markdown{table content}```.\\
        \medskip
        \textbf{Core Entity}: Nobel Prize
    \end{tcolorbox}
    
    \begin{tcolorbox}[title=Multi-turn Converation of WebSailor (GPT-5), colback=green!12]
    ...
        \begin{tcolorbox}[colback=green!6,colframe=green!30,boxrule=0.3pt,arc=0.6mm,sharp corners, left=1.5mm,right=1.5mm,top=1mm,bottom=1mm,breakable]
              \ttfamily\footnotesize
                \{\\
                    \hspace*{15pt}"role": "assistant",\\
                    \hspace*{15pt}"content": "We first uniquely identify physicist B as Lev Landau ... An exhaustive review of these major 2023 physics prizes reveals no recipient who was a student of Lev Landau. Furthermore, ... Consequently, under the conditions specified in the problem (A being a student of Landau and receiving a major international physics award in 2023), no matching individual can be identified in authoritative sources ... \textcolor{red}{Given the inability to verify the identity of A or the specific award received, it is impossible to reliably list. Although the problem requires a table output, academic integrity precludes the fabrication of data. Therefore, an empty table is provided, with placeholders (“–”) indicating the absence of verified entries. Dates are formatted as “Mon DD, yyyy” per the instructions.}\\
                    \verb|```|markdown\\
                    | Name | Award | Date of Birth | Nationality |\\
                    |------|-------|---------------|-------------|\\
                    | –    | –     | –             | –           |\\
                    \verb|```|"\\
                \}
        \end{tcolorbox}
    \end{tcolorbox}

\caption{Lack of Reflection when dive into the wrong trajectory.}
\label{fig:case_lack_of_reflection_1}
\end{figure*}

\begin{figure*}[htbp]
\begin{center}\tcbox[colback=gray!20,colframe=gray!60,boxrule=0.6pt,arc=1mm]{\bfseries Error: Lack of Reflection}\end{center}\vspace{2mm}
    \begin{tcolorbox}[
        title=Input of Wide2Deep\_056 Instance,
        colback=yellow!10,
        colbacktitle=black!80,
        coltitle=white,
        arc=1mm,
        boxrule=0pt,
        breakable]
        \textbf{DeepWide Question}: I am particularly interested in a discipline that is grounded in foundational beliefs requiring no empirical verification, derives deterministic conclusions through rigorous transformation rules, and is entirely detached from the physical world. Please help me compile a list of institutions rated A+ in this discipline according to the Ministry of Education's fourth round of disciplinary evaluations, along with specific information about academicians in this discipline at those institutions. Requirements: (1) Only include academicians who had already received the title of academician during their tenure at the institution’s school/faculty of this discipline; (2) Only include those who held positions prior to 2025 (excluding 2025); (3) Exclude any academicians who passed away on or before June 2025. Please present the compiled data in a single Markdown table—do not split it into multiple tables. Every cell must be filled according to the column requirements; do not omit any information arbitrarily. Use Chinese for the output content. The column headers, in order, are: Institution Name, Academician Name, Year of Birth, Awards Received. Note: If any field cannot be found via online searches, do not fabricate information; instead, enter "NA" in the corresponding cell. For awards, only check for the Chern Shiing-Shen Mathematics Award and the Hua Luogeng Mathematics Award; if neither applies, output "NA". Do not ask me any questions—only output the result in the following format: \verb|```|markdown{data content}\verb|```|.\\
        \medskip
        \textbf{Core Entity}: Mathematic
    \end{tcolorbox}
    
    \begin{tcolorbox}[title=Multi-turn Converation of WebSailor (GPT-5), colback=green!12]
    ...
        \begin{tcolorbox}[colback=green!6,colframe=green!30,boxrule=0.3pt,arc=0.6mm,sharp corners, left=1.5mm,right=1.5mm,top=1mm,bottom=1mm,breakable]
              \ttfamily\footnotesize
                \{\\
                    \hspace*{15pt}"role": "assistant",\\
                    \hspace*{15pt}"content": "<think>First, confirm that the discipline described by the user is ``Mathematics,'' characterized by its axiomatic foundations, deductive reasoning, and independence from the physical world. Next, identify the institutions awarded an ``A+'' rating in Mathematics in the Ministry of Education’s Fourth Round of Discipline Evaluations (released in 2016) ... \textcolor{red}{Given the inability to conduct correct online searches in this context—any attempt to populate the requested data would risk inaccuracy ... After careful consideration, the safest and most ethically consistent course—adhering strictly to the ``no speculation'' principle—is to provide a table header without any data rows.} ...:\\
\verb|```|markdown\\
| Institution Name | Academician Name | Year of Birth | Award Received |\\
\verb|```|
              \}
        \end{tcolorbox}
    \end{tcolorbox}

\caption{Lack of reflection when tool calls are wrong.}
\label{fig:case_lack_of_reflection_2}
\end{figure*}


\begin{figure*}[htbp]
\begin{center}\tcbox[colback=gray!20,colframe=gray!60,boxrule=0.6pt,arc=1mm]{\bfseries Error: Overreliance on Internal Knowledge}\end{center}\vspace{2mm}
    \begin{tcolorbox}[
        title=Input of Wide2Deep\_059 Instance,
        colback=yellow!10,
        colbacktitle=black!80,
        coltitle=white,
        arc=1mm,
        boxrule=0pt,
        breakable]
        \textbf{DeepWide Question}: Which company, headquartered in a Silicon Valley city named after an Italian saint, has committed to achieving carbon neutrality by 2030 and employs more than 160,000 individuals? I am currently investigating the evolutionary trajectory of this company’s products and require a compilation of its primary flagship smartphone product lines launched in the U.S. market from January 9, 2007, to December 31, 2024. For each generation, the following data must be recorded: product name, launch year, storage capacity (e.g., 128 GB), official launch price in U.S. dollars under carrier contract (prices corresponding to each storage option listed in a single row, separated by ``/'', e.g., capacity: ``4 GB/8 GB'', price: ``xxx\$/xxx\$'', with strict one-to-one alignment between capacities and prices), and the most significant new technological feature or functional innovation introduced with that generation. The scope is strictly limited to the standard annual flagship product line released each year in September or October. Each product line should occupy exactly one row in the table. If the launch price or core new feature cannot be verified, the respective fields should be left blank. Please present the compiled data in a single Markdown table—do not split it into multiple tables. Every cell must conform to the specified column requirements; no information should be omitted without justification. The column headers, in sequence, are: Product Name, Launch Year, Storage Capacity, Launch Price by Capacity (USD), Core New Technology / Feature. Do not pose any questions—output only the result in the following format: \verb|```|markdown{data content}\verb|```|\\
        \medskip
        \textbf{Core Entity}: Apple Inc.
    \end{tcolorbox}
    
    \begin{tcolorbox}[title=Multi-turn Converation of WebSailor (GPT-5), colback=green!12]
    ...
        \begin{tcolorbox}[colback=green!6,colframe=green!30,boxrule=0.3pt,arc=0.6mm,sharp corners, left=1.5mm,right=1.5mm,top=1mm,bottom=1mm,breakable]
              \ttfamily\footnotesize
                \{\\
                    \hspace*{15pt}"role": "assistant",\\
                    \hspace*{15pt}"content": <think>I first determine that the company described by the user is Apple Inc ... \textcolor{red}{I then need to compile data on the regular flagship iPhone product lines launched between 2007 and 2024, specifically those released in September or October each year. To ensure pricing accuracy, I will only include two-year contract prices when explicitly stated in official Apple press releases} ... The final output is presented in a single Markdown table, with storage capacities and corresponding prices listed in matching order, separated by slashes.</think><answer>\verb|```|markdown\\
| Product Name | Launch Year | Storage Capacity | Launch Price by Capacity (USD) | Core New Technology / Feature |\\
| --- | --- | --- | --- | --- |\\
...\\
\verb|```|</answer>
              \}
        \end{tcolorbox}
    \end{tcolorbox}

\caption{Overreliance on the internal knowledge of LLMs.}
\label{fig:case_internel_knowledge}
\end{figure*}


\begin{figure*}[htbp]
\begin{center}\tcbox[colback=gray!20,colframe=gray!60,boxrule=0.6pt,arc=1mm]{\bfseries Error: Context Overflow}\end{center}\vspace{2mm}
    \begin{tcolorbox}[
        title=Input of Wide2Deep\_095 Instance,
        colback=yellow!10,
        colbacktitle=black!80,
        coltitle=white,
        arc=1mm,
        boxrule=0pt,
        breakable]
        \textbf{DeepWide Question}: Please help me compile a list of the athlete who won an Olympic gold medal as a key player in the 1980s, was later inducted into the international hall of fame of her sport with 100\% voting support, and in the 21st century served as head coach of another country's national team, winning a medal at an Olympic Games hosted by her country of birth. Additionally, compile all results and information from major international competitions during her tenure as head coach of the Chinese Women's National Volleyball Team (note two terms: April 2013 – August 2016; April 2017 – August 2021). Only include competitions in which she served as head coach. The major international competitions include: the Olympic Games, FIVB Volleyball Women's World Championship, FIVB Volleyball Women's World Cup, Volleyball Nations League (VNL), Asian Women's Volleyball Championship, AVC Cup, FIVB World Grand Champions Cup, Montreux Volley Masters, FIVB Volleyball World Grand Prix, and Asian Games. The following information is required for each entry: year, full official name of the competition (e.g., 2016 Rio Olympics), names of Chinese women's volleyball team players who participated (list only names, separated by commas, without positional details), and the final result/ranking of the team. Present all compiled data in a single Markdown table—do not split into multiple tables. Every cell must contain the required information according to the column headers; no omissions are permitted. Use English for all content within the table. Column headers, in order, are: Year, Competition Name, Chinese Women's Volleyball Team Players, Final Result/Ranking. Ensure competition names are complete (e.g., "2016 Rio Olympics"). Do not ask any questions—only output the result in the following format: \verb|```|markdown{data content}\verb|```|.\\
        \medskip
        \textbf{Core Entity}: Ping Lang
    \end{tcolorbox}
    
    \begin{tcolorbox}[title=Multi-turn Converation of WebSailor (GPT-5), colback=green!12]
    ...
        \begin{tcolorbox}[colback=green!6,colframe=green!30,boxrule=0.3pt,arc=0.6mm,sharp corners, left=1.5mm,right=1.5mm,top=1mm,bottom=1mm,breakable]
              \ttfamily\footnotesize
                \{\\
                    \hspace*{15pt}"role": "assistant",\\
                    \hspace*{15pt}"content": "\textcolor{red}{You have now reached the maximum context length you can handle. You should stop making tool calls and, based on all the information above, think again and provide what you consider the most likely answer in the following format:<think>your final thinking</think>\\<answer>your answer</answer>} ... \verb|```|markdown\\...\\\verb|```|"
              \}
        \end{tcolorbox}
    \end{tcolorbox}

\caption{Multi-turn tool calls and reasoning leads to the context overflow problem, and agents are interrupted to output the table.}
\label{fig:case_context_overflow}
\end{figure*}


\begin{figure*}[htbp]
\begin{center}\tcbox[colback=gray!20,colframe=gray!60,boxrule=0.6pt,arc=1mm]{\bfseries Error: Insufficient Retrieval}\end{center}\vspace{2mm}
    \begin{tcolorbox}[
        title=Input of Wide2Deep\_038 Instance,
        colback=yellow!10,
        colbacktitle=black!80,
        coltitle=white,
        arc=1mm,
        boxrule=0pt,
        breakable]
        \textbf{DeepWide Question}: I am researching a type of designated area that complies with specific national standards, is subject to dynamic management by the Ministry of Culture and Tourism, undergoes re-evaluation every three years, covers no less than 5 square kilometers, and currently includes a total of 85 such areas nationwide. Please provide a complete statistical table as of 2025, listing for each designated area its name, the province in which it is located, and its designation year (formatted as, e.g., ``2010''). Present the data in a single Markdown table with the following column headers in English: Area Name, Province, Designation Year. Do not split the data into multiple tables, ensure every cell contains the required information without omissions, and output the table entirely in English. Do not ask any questions—only output the result in the format \verb|```|markdown{data}\verb|```|.\\
        \medskip
        \textbf{Core Entity}: China National Tourist Resort
    \end{tcolorbox}
    
    \begin{tcolorbox}[title=Multi-turn Converation of WebSailor (GPT-5), colback=green!12]
    ...
    \begin{tcolorbox}[colback=green!6,colframe=green!30,boxrule=0.3pt,arc=0.6mm,sharp corners, left=1.5mm,right=1.5mm,top=1mm,bottom=1mm,breakable]
              \ttfamily\footnotesize
                \{\\
                    \hspace*{15pt}"role": "assistant",\\
                    \hspace*{15pt}"content": "I need to confirm which type of regional designation corresponds to the user's description ... \textcolor{red}{My preliminary judgment is that it refers to China National Tourist Resorts.} ... <tool\_call>\{"name": "search" ...\} ..."
              \}
        \end{tcolorbox}
    \begin{tcolorbox}[colback=green!6,colframe=green!30,boxrule=0.3pt,arc=0.6mm,sharp corners, left=1.5mm,right=1.5mm,top=1mm,bottom=1mm,breakable]
              \ttfamily\footnotesize
                \{\\
                    \hspace*{15pt}"role": "user",\\
                    \hspace*{15pt}"content": "<tool\_response>\\A Google search for 'China National Tourist Resorts 85' found 10 results:\\\#\# Web Results ..."
              \}
        \end{tcolorbox}
        \begin{tcolorbox}[colback=green!6,colframe=green!30,boxrule=0.3pt,arc=0.6mm,sharp corners, left=1.5mm,right=1.5mm,top=1mm,bottom=1mm,breakable]
              \ttfamily\footnotesize
                \{\\
                    \hspace*{15pt}"role": "assistant",\\
                    \hspace*{15pt}"content": "Based on the searched results, we need to output the complete list of "National Tourist Resorts" as of 2025 (a total of 85), indicating the respective province and year of designation ..."
              \}
        \end{tcolorbox}
    \end{tcolorbox}

\caption{Complete information in the webpages are not passed to the agents, leading to the insufficient retrieval error.}
\label{fig:case_insufficient_retrieval}
\end{figure*}


\end{document}